\RequirePackage{fix-cm}
\documentclass[twocolumn]{svjour3}          
\smartqed  
\usepackage{graphicx}
\usepackage{natbib}
\usepackage{float} 

\usepackage{amsmath} 
\usepackage{mathtools}
\usepackage{amsfonts}
\usepackage{amssymb}

\makeatletter 
\@fleqnfalse
\@mathmargin\@centering
\makeatother

\usepackage{booktabs} 
\usepackage[normalem]{ulem}
\usepackage{xcolor, colortbl} 

\usepackage[colorlinks=true,linkcolor=blue]{hyperref}%

\newcommand{\mat}[1]{\textnormal{\textbf{#1}}}
\newcommand{\spd}[1]{\textnormal{SPD}(#1)}
\newcommand{\minorrev}[1]{\textcolor{black}{#1}}
\newcommand{\majorrev}[1]{{#1}}

\usepackage{caption} 

\journalname{Brain Imaging and Behavior}
\begin{document} 

\title{Predicting cognitive scores with graph neural networks through sample selection learning 
}

\titlerunning{Sample selection for IQ prediction on connectomes using GNNs}        

\author{Martin Hanik\thanks{Martin Hanik and Mehmet Arif Demirta\c{s} contributed equally to this work.} \and
        Mehmet Arif Demirta\c{s} \and
        Mohammed Amine Gharsallaoui \and
        Islem Rekik\thanks{corresponding author: irekik@itu.edu.tr, \url{http://basira-lab.com/}}
}


\institute{M. Hanik \at
              Zuse Institute Berlin, Berlin, Germany
            \and
            M. A. Demirta\c{s} \at
                BASIRA Lab, Faculty of Computer and Informatics, Istanbul
                Technical University, Istanbul, Turkey
            \and
            M. A. Gharsallaoui \at
            	Ecole Polytechnique de Tunisie (EPT), Tunis, Tunisia; 
				BASIRA Lab, Faculty of Computer and Informatics, Istanbul
            Technical University, Istanbul, Turkey
			\and
			I. Rekik \at 
				BASIRA Lab, Faculty of Computer and Informatics, Istanbul Technical University, Istanbul, Turkey;
				Computing, School of Science and Engineering, University of Dundee, Dundee, UK
}

\date{Received: date / Accepted: date}

\maketitle

\begin{abstract}
Analyzing the relation between intelligence and neural activity is of the utmost importance in understanding the working principles of the human brain in health and disease. In existing literature, functional brain connectomes have been used successfully to predict cognitive measures such as intelligence quotient (IQ) scores in both healthy and disordered cohorts using machine learning models. However, existing methods resort to flattening the brain connectome (i.e., graph) through vectorization which overlooks its topological properties. To address this limitation and inspired from the emerging graph neural networks (GNNs), we design a novel regression GNN model (namely RegGNN) for predicting IQ scores from brain connectivity. On top of that, we introduce a novel, fully modular sample selection method to select the best samples to learn from for our target prediction task. However, since such deep learning architectures are computationally expensive to train, we further propose a \emph{learning-based sample selection} method that learns how to choose the training samples with the highest expected predictive power on unseen samples. For this, we capitalize on the fact that connectomes (i.e., their adjacency matrices) lie in the symmetric positive definite (SPD) matrix cone. Our results on full-scale and verbal IQ prediction outperforms comparison methods in autism spectrum disorder cohorts and achieves a competitive performance for neurotypical subjects using 3-fold cross-validation. Furthermore, we show that our sample selection approach generalizes to other learning-based methods, which shows its usefulness beyond our GNN architecture.


\end{abstract}

\section{Introduction} \label{intro}

Understanding how the structure of the brain influences cognitive scores such as IQ plays a vital role in understanding the working principles of the human brain. Cognitive scores are indicators of intellectual capacity which were found to be strongly connected to social factors: while high correlation between intelligence scores measured in childhood and educational success were observed in~\citep{Colom:2007,Deary:2007}, they were also linked to health and mortality~\citep{Gottfredson:2004,Batty:2007}. Motivated by this fact, many studies have investigated how far intelligence quotients (IQ) can be predicted
from the structure of the brain. It was found, for example, that cerebral volume positively correlates with cognitive ability~\citep{Reiss:1996,Mcdaniel:2005}. On a finer scale, activity and global connectivity of parts of the brain, especially of the lateral prefrontal cortex, are linked to IQ~\citep{Gray:2003,Woolgar:2010,Cole:2012,Cole:2015}.

Against this background, recent works have explored the possibility to predict cognitive ability scores from functional brain connectomes~\citep{Pamplona:2015,Dubois:2018,Dadi:2019,Dryburgh:2020,HE:2020,Jiang:2020}. Conventionally, connectomes are obtained from resting-state MRI and characterize the network structure of the brain; they are modeled as graphs whose nodes represent regions of interest (ROIs) and whose edges correspond to correlations in activity between these ROIs~\citep{Sporns:2005}. In order to achieve better generalizability across contexts and populations, \citep{Shen:2017} proposed a data-driven protocol for connectome-based predictive modeling of brain-behavior relationships, using cross-validation, to train a linear regression model. Building upon it, \citep{Dryburgh:2020} improved the results by evaluating negative and positive correlations of brain regions separately. They performed their analysis on both neurotypical subjects and subjects with Autism Spectrum Disorder (ASD) in order to investigate how neural correlates of intelligence scores are altered by atypical neurodevelopmental disorders.

Although such works achieved significant success, they mainly relied on classical machine learning approaches, which do not incorporate the \emph{graph structure} of the connectomes; therefore, the local and global topological properties of the connectomes are not leveraged. \citep{HE:2020} introduced graph neural networks (GNNs)~\citep{Wu:2021}, a subfield of \emph{geometric deep learning}, where learning is customized to non-Euclidean spaces such as graphs with complex topologies~\citep{Dehmamy:2019}. GNNs are deep neural networks with graph convolution layers. They have already lead to significant increases in performance over existing methods in many fields. For example, they have been successfully applied to classification tasks on networks~\citep{Kipf:2017,Qu:2019}, image segmentation~\citep{Qi:2017}, feature matching~\citep{Sarlin:2020}, few-shot learning~\citep{Garcia:2017,Kim:2019}, and various graph mining tasks \citep{Schlichtkrull:2018,Yun:2019,Zhang:2019}. A very recent review on GNNs in the field of network neuroscience \citep{Bessadok:2021} examined a variety of graph-based architecture tailored for brain connectivity classification, integration, superresolution and synthesis across time and modalities. However, none of the reviewed methods were designed for brain graph regression for cognitive score prediction.

In this paper, we propose the first GNN architecture, namely \emph{RegGNN}, that is specialized in regressing brain connectomes to a target cognitive score to predict. Our GNN utilizes graph convolutional layers to map input connectomes onto their corresponding cognitive scores, thereby allowing to extract the learned weights to identify the brain connectivities between anatomical regions that fingerprint the target score.  

To improve the performance of the GNN, we additionally propose a novel \emph{learning-based sample selection} method. It is independent from RegGNN and can be used with any architecture or regression learner. The method identifies training samples with the highest predictive power (i.e., those that are most likely to predict unseen test subjects with the lowest error); only these are then used for training. Through this, we eliminate the samples that do not increase---or even decrease---the prediction success of the model and reduce the computational resources needed for training the GNN. 

Within our sample selection method, we make use of the fact that the (weighted) adjacency matrix of a functional brain connectome, when modeled as a correlation matrix, is symmetric positive semi-definite; and becomes symmetric positive definite after a simple regularization step~\citep{Dodero:2015,Wong:2018, You:2020}.
The space of SPD matrices forms a nonlinear manifold~\citep{Arsigny:2006}, and like \citep{You:2020}, we use a Riemannian geometric structure on it in order to obtain a \emph{natural} notion of distance between two connectomes as well as tangent matrices that encode the paths that realize this distance. 


We summarize the main contributions of our work as follows: 
\begin{enumerate}
    \item We introduce a novel, learning-based sample selection method for graph neural networks that helps to increase accuracy when predicting cognitive scores from connectomes.
    \item We propose novel similarity measures between brain connectomes by combining notions from Riemannian geometry and topology of graphs. These measures can be used in other applications whenever we deal with objects that can be interpreted as elements of Riemannian manifolds.
    \item We design a pipeline, consisting of RegGNN with sample selection, which outperforms state-of-the-art models in predicting full scale intelligence and verbal intelligence quotients from functional brain connectomes in an autism spectrum disorder cohort and achieves a competitive performance in a neurotypical cohort. 
\end{enumerate}

\section{Methods}
In this section, we detail the architecture of our RegGNN. Furthermore, we introduce our proposed sample selection method and show how we incorporate it into the training process of the GNN. To start with, we recount some facts on the Riemannian geometry of SPD matrices. Furthermore, we recall graph-topological centrality measures. The mathematical notations that we use in the following are summarized in Table~\ref{tab-notations}.

\begin{table*}[ht]
\captionsetup{justification=centering}
\centering
\begin{scriptsize}
\begin{tabular}{ >{\centering\arraybackslash}m{.15\textwidth} >{\centering\arraybackslash}m{.15\textwidth} >{\centering\arraybackslash}m{.65\textwidth} }
\toprule
	Notation& 
	Dimension& 
	Definition \\
	\toprule
    $n_{\text{Train}}$ & $\mathbb{N}$ & number of subjects in the training group \\
    $n_{\text{Test}}$ & $\mathbb{N}$ & number of subjects in the test group \\
	$d$ & $\mathbb{N}$ & number of brain regions (i.e, ROIs) \\
	$n_s$ & $\mathbb{N}$ & number of elements in the train-in group \\
	$n_h$ & $\mathbb{N}$ & number of elements in the holdout group \\
	$\textnormal{SPD}(n)$ & - & manifold of $n \times n$ symmetric positive definite matrices\\
	$T_{\textnormal{\textbf{P}}}\textnormal{SPD}(n)$ & - & tangent space at $\textnormal{\textbf{P}} \in \textnormal{SPD}(n)$ (can be identified with the vector space of symmetric matrices of the same size)\\
	$\textnormal{\textbf{I}}$ & $\mathbb{R}^{d \times d}$ & identity matrix\\
	$\textnormal{\textbf{P}}^s_i$ & $\mathbb{R}^{d \times d}$ & functional brain connectome (SPD) of subject $i$ from the train-in group $s$ \\
	$\textnormal{\textbf{P}}^h_l$ & $\mathbb{R}^{d \times d}$ & functional brain connectome (SPD) of subject $l$ from the holdout group $h$\\
	$\tilde{\textnormal{\textbf{S}}}^{s,s}_{i,j}$ & $\mathbb{R}^{d \times d}$ & tangent matrix (symmetric matrix) encoding the geodesic between connectomes $i$ and $j$ from the train-in group s at $\textnormal{\textbf{P}}^s_i$\\
	$\tilde{\textnormal{\textbf{S}}}^{s,h}_{j,l}$ & $\mathbb{R}^{d \times d}$ & tangent matrix (symmetric matrix) encoding the geodesic between connectomes $j$ from the train-in group $s$ and $l$ from the holdout group $h$ at $\textnormal{\textbf{P}}^s_j$\\
	$\textnormal{\textbf{S}}^{s,s}_{i,j}$ & $\mathbb{R}^{d \times d}$ & parallel translation (symmetric matrix) of  $\tilde{\textnormal{\textbf{S}}}^{s,s}_{i,j}$ to $T_\textnormal{\textbf{I}}\textnormal{SPD}(n)$\\ 
    $\textnormal{\textbf{S}}^{s,h}_{j,l}$ & $\mathbb{R}^{d \times d}$ & parallel translation (symmetric matrix) of  $\tilde{\textnormal{\textbf{S}}}^{s,h}_{j,l}$ to $T_\textnormal{\textbf{I}}\textnormal{SPD}(n)$\\ 
    $v^{s,s}_{i,j}$ & $\mathbb{R}^{d}$ & feature vector extracted from $\textnormal{\textbf{S}}^{s,s}_{i,j}$\\
    $v^{s,h}_{j,l}$ & $\mathbb{R}^{d}$ & feature vector extracted from $\textnormal{\textbf{S}}^{s,h}_{j,l}$\\
    $IQ^{s}_{i}$     & $\mathbb{R}$ & cognitive score (IQ) of subject $i$ from the train-in group $s$\\
    $IQ^{h}_{l}$     & $\mathbb{R}$ & cognitive score (IQ) of subject $l$ from the holdout group $h$\\

\bottomrule
\end{tabular}
\end{scriptsize}
\caption{Major mathematical notations used in this paper.}
\label{tab-notations}
\end{table*}

\subsection{Preliminaries}

    The space of $n$-by-$n$ symmetric positive definite matrices $\textnormal{SPD}(n) = \{\textnormal{\textbf{P}} \in \mathbb{R}^{n,n}: \textnormal{\textbf{P}}^T=\textnormal{\textbf{P}}, \text{all eigenvalues of \textbf{P}}$ $\text{are positive}\}$ forms a cone-like manifold in the set of all matrices of the same size~\citep{Faraut:1994}. Being a manifold, there is a well-defined tangent space at every point $\textnormal{\textbf{P}} \in \textnormal{SPD}(n)$, which we denote by $T_{\textnormal{\textbf{P}}}\textnormal{SPD}(n)$. It is a basic fact that each $T_{\textnormal{\textbf{P}}}\textnormal{SPD}(n)$ can be identified with the set of symmetric $n$-by-$n$ matrices. Therefore, in order to avoid later confusion, we call their elements \emph{tangent matrices} instead of ``tangent vectors'' (which is the standard term in differential geometry). 
    
    As a manifold, $\textnormal{SPD}(n)$ can be endowed with a Riemannian geometric structure~\citep{doCarmo:1992}. Such a structure is determined by the choice of a Riemannian metric, i.e., a smoothly varying inner product on the tangent spaces. With its help, we can measure angles between (and norms of) tangent matrices. Furthermore, it induces a distance $d$ on the space. Consequently, geodesics can be defined as (locally) shortest paths. 
    Like a straight line in Euclidean space, a geodesic $\gamma$ that connects two points $\textnormal{\textbf{P}}, \textnormal{\textbf{Q}} \in \textnormal{SPD}(n)$ can be represented by a unique tangent matrix $\log(\textnormal{\textbf{P}}, \textnormal{\textbf{Q}}) \in T_{\textnormal{\textbf{P}}}\textnormal{SPD}(n)$.\footnote{It is denoted like this because the corresponding map is called \emph{Riemannian logarithm}.} In particular, $\log(\textnormal{\textbf{P}}, \textnormal{\textbf{Q}})$ points in the direction of $\textnormal{\textbf{Q}}$, i.e., is parallel to $\gamma$ at $\textnormal{\textbf{P}}$ and has norm (measured in the one induced from the metric) equal to the distance between $\textnormal{\textbf{P}}$ and $\textnormal{\textbf{Q}}$.
    Because of this, we can view $\log(\textnormal{\textbf{P}}, \textnormal{\textbf{Q}})$ as the linearized ``difference'' between $\textnormal{\textbf{Q}}$ and $\textnormal{\textbf{P}}$.
    
\begin{figure}[ht]
        \centering
        \includegraphics[width=.45\textwidth]{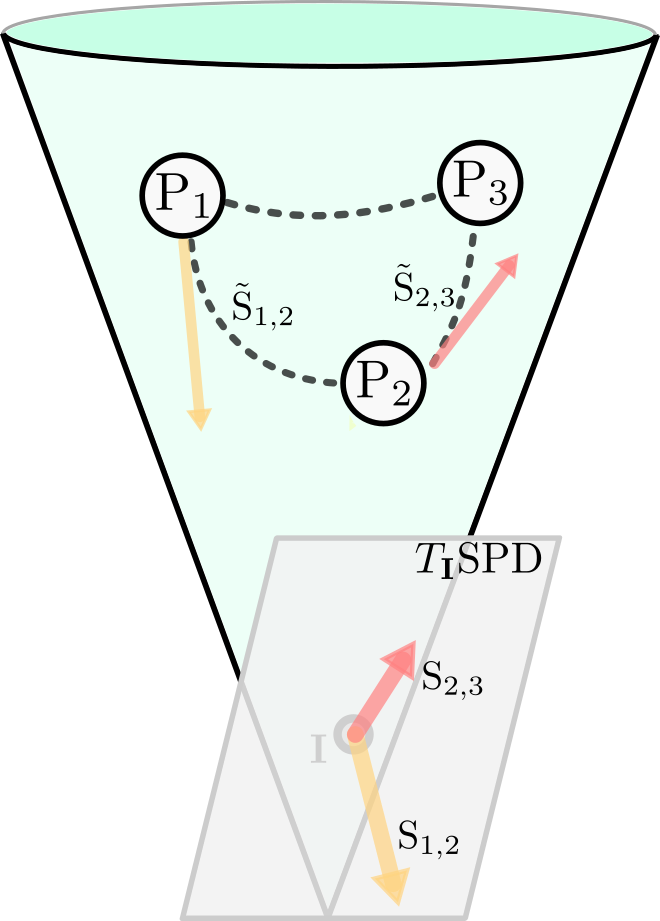}
        \caption{Illustration of geodesics and parallel transport of tangent matrices on the SPD cone. The dashed lines are the geodesics between the matrices $\mathbf{P}_1,\mathbf{P}_2,\mathbf{P}_3 \in \spd{n}$. The tangent matrices $\tilde{\mathbf{S}}_{1,2} : = \log(\mathbf{P}_1,\mathbf{P}_2)$ and $\tilde{\mathbf{S}}_{2,3} : = \log(\mathbf{P}_2,\mathbf{P}_3)$ are the yellow and red arrow, respectively; their parallel translations to \minorrev{the tangent space $T_{\mathbf{I}}\textnormal{SPD}$ at the identity matrix $\mathbf{I}$} are $\mathbf{S}_{1,2}$ and $\mathbf{S}_{2,3}$.}  
         
        \label{fig:cone}
    \end{figure}

    In contrast to Euclidean geometry, tangent matrices from different tangent spaces of a Riemannian manifold cannot be compared directly. Instead, they must be transported along curves to the same tangent space; 
    this process is called \emph{parallel translation}. 
    This means that although tangent matrices at different points $\textnormal{\textbf{P}} \in \textnormal{SPD}(n)$ and $\textnormal{\textbf{Q}} \in \textnormal{SPD}(n)$ are symmetric matrices, we must bring them to a common point in order to compare them. 
    The SPD space and parallel translation of vectors are illustrated in Fig.~\ref{fig:cone}.
    
    Since all notions depend on the Riemannian structure, we must fix one. 
    \minorrev{For $\textnormal{SPD}(n)$,} several can be found in the literature, the most popular being the Log-Euclidean metric~\citep{Arsigny:2006} and the affine-invariant metric~\citep{Moakher:2005,Pennec:2006}.
    They have been applied successfully to connectomes for classification~\citep{Dodero:2015,Yamin:2020}, regression~\citep{Wong:2018}, fingerprint extraction~\citep{Abbas:2021}, and statistical analysis~\citep{You:2020}. We choose to work with the Log-Euclidean metric because it allows for comparatively efficient algorithms. Furthermore, parallel transport does not depend on the chosen path and a unique length-minimizing geodesic exists between any two points---both properties do not hold for most other metrics.
    \bigskip
    
    We now recall three basic topological centrality measures for an undirected\footnote{Of course, the centrality measures can also be defined for directed graphs but we do not need this here.} graph $G$: degree centrality, eigenvector centrality, and closeness centrality; we recount them in Appendix~\ref{Sec:AppA}.
    \minorrev{They measure how far} a node is central to the (graph) network in the sense that most of the communication passes through it.
    A good reference on this is the book~\citep{Fornito:2016}.
    \bigskip
    
    We are now ready to introduce the graph neural network, and afterwards, the sample selection process.

\subsection{RegGNN}

    \begin{figure*}[ht]
    \centering
    \includegraphics[width=\textwidth]{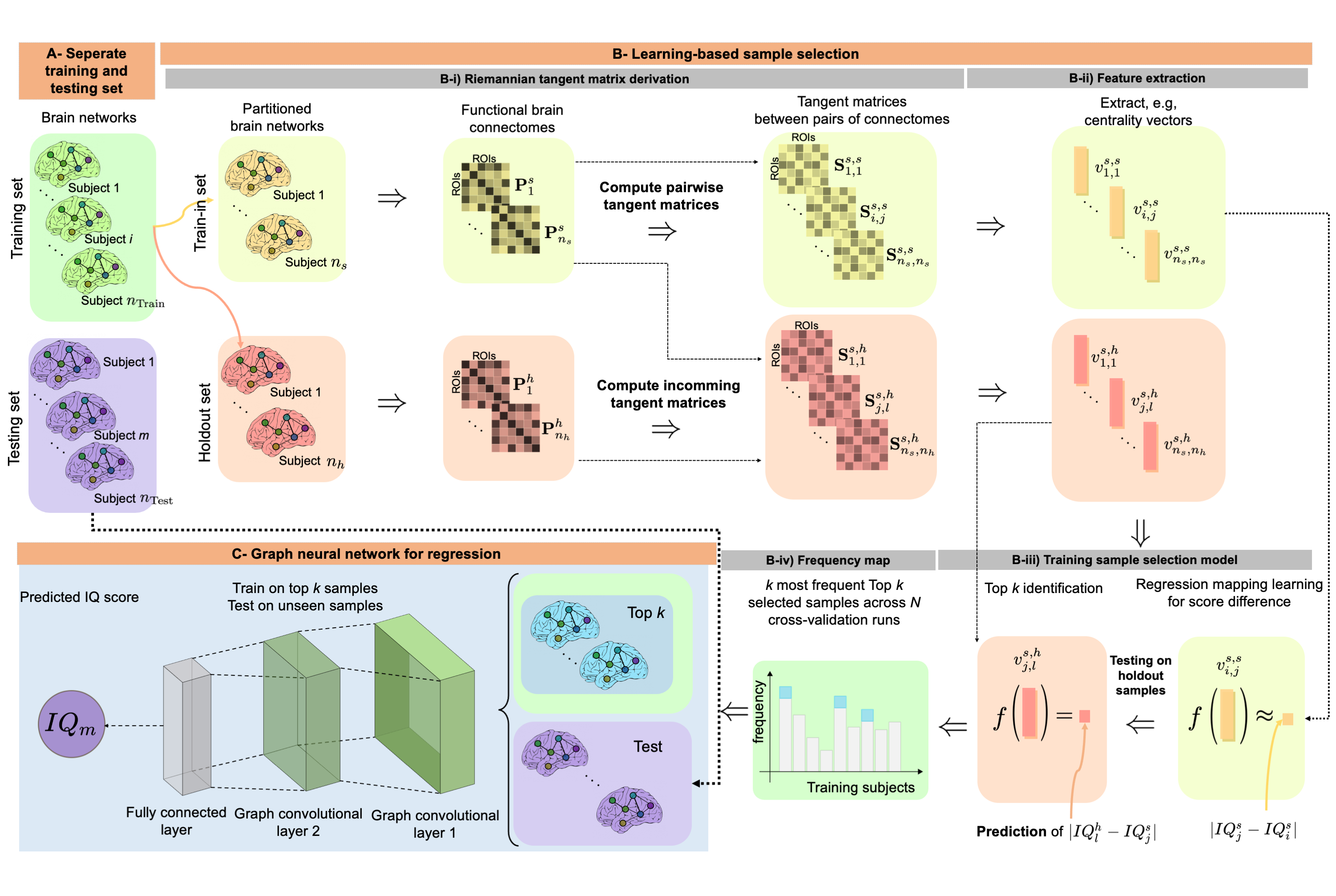}
    \caption{\minorrev{\emph{Illustration of the proposed sample selection strategy to train our regression graph neural network RegGNN.} \textbf{A)} We split the data in training (green) and testing (violet) sets. \textbf{B-i)} After the training set is divided into a train-in (yellow) and a holdout (red) set, we extract tangent matrices for geodesics connecting elements from the train-in set (yellow) and tangent matrices encoding geodesics from elements of the train-in to elements of the holdout group (red).  \textbf{B-ii)} The information from the tangent matrices is compressed into vectors through topological feature extraction. \textbf{B-iii)} With linear regression we train a mapping $f$ on the train-in-to-train-in feature vectors (yellow) to learn differences in target score and record for each element $j$ of the train-in group the $k$ elements from the holdout group for whom the predicted difference in target score to $j$ was smallest. \textbf{B-iv)} For each sample in the holdout set, we count how often it was among the top $k$ predictors for a sample from the train-in group. \textbf{C)} After repeating B) in an N-fold cross validation manner, the $k$ samples (blue) with the highest accumulated top-$k$ frequency are selected. \majorrev{After negative correlations have been set to zero} the graph neural network is trained (only) on them. Finally, the testing set is used to evaluate the overall performance. }
    } 
    \label{fig:main_figure}
    \end{figure*}
    

    Our GNN for regression, \textit{RegGNN}, consists of two graph convolution layers and a downstream fully connected layer; a visualization is on the bottom left of Figure~\ref{fig:main_figure}. 
    In the following we denote the number of ROIs by $d$. Since adjacency matrices of connectomes are $d$-by-$d$ correlation matrices $\mathbf{C}$, they can have zero (but no negative) eigenvalues. Therefore, we can simply regularize them to being symmetric positive definite by adding a small multiple of the identity matrix $\mathbf{I}$, i.e.,
    \begin{equation} \label{eq:regularize}
       \mathbf{P} := \mathbf{C} + \mu \mathbf{I}
    \end{equation}
    for some small $\mu > 0$;
    see \citep{Dodero:2015} or \citep{Wong:2018}.
    \majorrev{For training RegGNN---but not for the sample selection---we set all negative eigenvalues to zero, as positive correlations have been shown to be more important in brain network analysis~\citep{Fornito:2016}. Indeed, in our experiments the results improved when negative correlations were ignored.}
    RegGNN receives the regularized, positive adjacency matrix $\mathbf{P}$ of a connectome and predicts the corresponding IQ score from it by applying graph convolutions.
    
    In the literature, there are various implementations of graph convolutions, which mainly differ by the propagation rule. For $i=0,1,2$ let $\mat{H}^{(i)}$ denote the activation matrix at the $i$-th layer with initialization $\mat{H}^{(0)} := \mat{I}$. It is propagated to the next layer according to the general rule $\mat{H}^{(i+1)} = g_i(\mat{H}^{(i)}, \mat{P})$ with functions $g_i: \mathbb{R}^{d,d_i} \times \spd{d} \to \mathbb{R}^{d,d_{i+1}}$ for $i=0,1$, where $d_0 := d$ and $d_2 := 1$. (The integer $d_1$ can be chosen appropriately---we use a suitable number near $d/2$.)
    We choose the $g_i$ as proposed in \citep{Kipf:2017}.
    Define $\tilde{\mathbf{P}} := \mathbf{P} + \mathbf{I}$ and let $\tilde{\mathbf{D}}$ be the diagonal degree matrix of $\tilde{\mathbf{P}}$, we then formalize $g_i$ as follows:
 
    $$g_i(\mathbf{H}^{(i)},\mathbf{P}) := \textnormal{ReLU} (\tilde{\mathbf{D}}^{-\frac{1}{2}} \tilde{\mathbf{P}} \tilde{\mathbf{D}}^{-\frac{1}{2}} \mathbf{H}^{(i)} \mathbf{W}^{(i)}),$$
    where $\mathbf{W}^{(i)} \in \mathbb{R}^{d_i,d_i}$ is the learnable weight matrix.
    We thus use the graph convolution layers to reduce the size of the connectomes and obtain an embedding for the brain graphs into $\mathbb{R}^d$. We apply a dropout layer after the first graph convolution operation for regularization. Finally, the obtained embedding passes through a fully connected layer (linear layer) which produces a continuous scalar output. The goal of the linear layer is to embed the resulting vector containing $d$ features into a scalar value representing the predicted IQ score. 

\subsection{Learning-based Sample Selection}
\label{Sec:sample_selection}
    We now introduce our \emph{learning-based} sample selection strategy. The underlying idea is the following. 
    Imagine the (rather extreme) case that our subjects are clustered (possibly with outliers) in $k$ tight groups according to their cognitive scores. Then, training a GNN on $k$ representatives\minorrev{, one from each cluster,} should yield good results; it should even perform better than a GNN trained on the full data set because it was not ``distracted'' by outliers during training. Ideally, as representatives we would choose the $k$ most central samples of each group, i.e., those with the smallest average \emph{difference in cognitive score} to the other samples. Now, since we want to predict cognitive scores from connectomes, we do not know the differences beforehand. 
    \majorrev{On the other hand, existing studies validated the relationship between brain connectivity patterns and brain behavior and cognition. For instance, recent papers~\citep{Pamplona:2015,Shen:2017,Dubois:2018,Dadi:2019,Dryburgh:2020,Jiang:2020,HE:2020} have shown that the cognitive ability of a person can be predicted quite accurately from the human connectome, indicating that the brain cognitive and behavior are encoded in its connectivity to a measurable degree.  Such prediction would have been elusive if similar data inputs (here brain connectomes) cannot be mapped to similar outputs (here cognitive scores). Consequently, we assume that similar brain connectivity networks are correlated in cognition whereas brain connectomes that vary in topological patterns might elicit different cognitive scores. Such hypothesis might seem somewhat reductionist as there are many other factors that contribute to molding and predicting brain cognition such as genetics and epigenetics~\citep{Goldberg:2004,Deary:2006,Reichenberg:2009}. However, such factors remain out of the scope of this study.}
    Therefore, our idea is to use the \emph{differences} between the connectomes to learn the \emph{differences} between the target scores in order to identify those ``representatives''.
    Our experiments below show that this idea of---to represent predicted local aggregations of data by (few) representatives and training only with them---generalizes well to real data.
    
    Implementing the idea, we represent differences between connectomes by tangent matrices and assume that the difference in IQ between two subjects depends linearly on (notions deduced from) the tangent matrix $\log(\textnormal{\textbf{P}}, \textnormal{\textbf{Q}})$ that encodes the geodesic between the corresponding connectomes $\textnormal{\textbf{P}}, \textnormal{\textbf{Q}} \in \textnormal{SPD}(d)$. This model is flexible, but at the same time allows for fast computations.
    Our sample selection method \emph{learns} this linear map, which we call $f$ in the following, via regression and uses it to identify the $k$ samples with the lowest predicted average difference in target score to all other samples. As motivated above, we assume that they are representative of the whole set but do not contain (most of) the outliers that hinder successful training of the GNN. The structure and terminology of our method are inspired by the work of \citep{Errica:2019}.
    
    The sample selection method consists of four steps, which are visualized in part B of Fig.~\ref{fig:main_figure}.
    Given a connectome data set, these are repeated in a nested $N$-fold cross-validation manner to make our selection of samples more robust. In cross-validation, we split the data set into two groups: a training subset which we call \emph{train-in group}, and a validation subset which we call \emph{holdout group}; we perform different train-in and holdout group splits so that each sample from the training set will be in the train-in group exactly $N-1$ times. We denote the (constant) sizes of the train-in and the holdout sets by $n_s$ and $n_h$, respectively.

    \textit{i) Riemannian tangent matrix derivation.} 
    For each pair of regularized connectomes $\textnormal{\textbf{P}}^s_i$, $\textnormal{\textbf{P}}^s_j \in \textnormal{SPD}(d)$ in the train-in group, we compute the tangent matrix
    $$\tilde{\textnormal{\textbf{S}}}^{s,s}_{i,j} := \log(\textnormal{\textbf{P}}^s_i, 
    \textnormal{\textbf{P}}^s_j) \in T_{\textnormal{\textbf{P}}^s_i}\textnormal{SPD}(d)$$ 
    that encodes the geodesic between them and parallel translate it to $T_{\textnormal{\textbf{I}}}\textnormal{SPD}(d)$; we denote the resulting symmetric $d$-by-$d$ matrix by $\textnormal{\textbf{S}}^{s,s}_{i,j}$. As a result, we obtain a set of $n_s\, (n_s - 1)/2$ tangent matrices in $T_{\textnormal{\textbf{I}}}\textnormal{SPD}(d)$ that represent the pairwise differences between the connectomes from the train-in group. Analogously, we get a tangent matrix $\textnormal{\textbf{S}}^{s,h}_{j,l} \in T_{\textnormal{\textbf{I}}}\textnormal{SPD}(d)$ for each pair with one sample $\textnormal{\textbf{P}}^s_i$ from the train-in and another sample $\textnormal{\textbf{P}}^h_l$ from the holdout group; this results in another set consisting of $n_{s} \, n_{h}$ tangent matrices. The latter are the \emph{outgoing} ``difference matrices'' \emph{from the train-in into the holdout set}.
  
    \textit{ii) Topological feature extraction from tangent matrices.} 
    The tangent matrices are still rather high dimensional, which leads to long computation times. Thus, we suggest to extract topological features in order to encode the information in more compact form.
    We select degree, closeness, and eigenvector centrality as well as combinations of them as our candidates for feature extraction. Note that in our case a tangent matrix represents the ``difference'' between two connectomes. The above features thus encode information on linearized \emph{changes} in node connectivity. To the best of our knowledge, this is the first time that these notions were used in conjunction.
    As a result, from all in-group tangent matrices $\textnormal{\textbf{S}}^{s,s}_{i,j}$ as well as outgoing tangent matrices $\textnormal{\textbf{S}}^{s,h}_{j,l}$ we obtain feature vectors $v^{s,s}_{i,j}$ and $v^{s,h}_{i,j}$, respectively.
  
    \textit{iii) Learning a linear regression mapping for predictive sample selection.} 
    We \emph{learn} the linear map $f$ via regression by training to map the vectors $v^{s,s}_{i,j}$ corresponding to samples \textit{i} and \textit{j} from the train-in group to the \emph{absolute difference in target score} $|IQ_j^s-IQ^s_i|$ between them. 
    We then apply the learned linear regression mapping $f$ to the vectors $v^{s,h}_{j,l}$ to predict the differences in target score between all samples $j$ from the train-in and samples $l$ from the holdout group.
    
  
    \textit{iv) Frequency map.}
    We record \emph{for each} holdout sample $\textnormal{\textbf{P}}^h_l$ the $k$ subjects from the train-in group with the smallest predicted difference under $f$ and increment a frequency map (i.e., a counter) that is initialized at the start of the sample selection process. The frequency value of a subject is then the number of times it was one of the top \textit{k} predictive samples. These frequencies give an approximated ranking whereby the top samples are closest to the largest number of other samples in (predicted) target score.
    \bigskip
    
    After the cross-validation is finished, we extract the top \textit{k} samples\footnote{Note that we could pick a different number here. We leave exploring possible other choices for future work.} with the highest cumulative frequencies. We expect these samples to have the highest representative power as they consistently predicted samples in different holdout groups with low error.
    
\subsection{Training Process} \label{sec:training}
    In the following, we explain how we integrate the sample selection method into the training process of RegGNN. The whole pipeline is shown in Fig.~\ref{fig:main_figure}.
    
    Given the data set of connectomes our proposed training pipeline consists of the following steps A-C. 
    
    \textbf{A- Training-test split.} First, we split the data set into a training and a test set. The test set is used \emph{only} for the final evaluation of RegGNN.
    
    \textbf{B- Learning-based sample selection.} Then, we select the top $k$ samples with the highest representative power from the training set by applying the sample selection method from Sec.~\ref{Sec:sample_selection}.
    
    \textbf{C- RegGNN architecture for regression.}
     Finally, we train RegGNN on the top \textit{k} samples using cross-validation to evaluate model generalizability against perturbations of training and testing data distributions. The final testing is done on the unseen test set.

\subsection{Data and Methodology}
We used the pipeline from Sec.~\ref{sec:training} to predict the full scale intelligence quotient (FIQ) and the verbal intelligence quotient (VIQ) from brain connectomes for both neurotypical (NT) subjects as well as subjects with autism spectrum disorder (ASD). In the following, we summarize these experiments.

\begin{table*}[ht]
\scriptsize
\centering
\begin{tabular}{ccc}
\hline
\multicolumn{1}{c}{\textbf{Method}}          & \multicolumn{1}{c}{\textbf{Mean MAE $\pm$ std (min, max)}}                     & \textbf{Mean RMSE $\pm$ std (min, max)}\\ \hline
\multicolumn{3}{c}{\cellcolor[HTML]{eaFFd9} \textbf{NT (FIQ)}}\\ \hline
\multicolumn{1}{c}{CPM                    }  & \multicolumn{1}{c}{\textcolor{blue}{\textbf{9.672}}} & \textbf{12.440}\\
\multicolumn{1}{c}{CPM (a)              }  & \multicolumn{1}{c}{ 11.383 $\pm$ 4.806 (9.589, 28.675) } & 13.972 $\pm$ 4.986 (\underline{12.075}, 31.911) \\
\multicolumn{1}{c}{CPM (g)              }  & \multicolumn{1}{c}{ 10.546 $\pm$ 1.407 (9.729, 14.989) } & 13.358 $\pm$ 1.831 (12.393, 19.358) \\
\multicolumn{1}{c}{CPM (tm)     }  & \multicolumn{1}{c}{ 10.245 $\pm$ 2.086 (9.523, 17.745) } & 12.959 $\pm$ 2.515 (12.096, 22.009) \\
\multicolumn{1}{c}{CPM (dc)             }  & \multicolumn{1}{c}{ 10.577 $\pm$ 2.487 (9.561, 19.394) } & 13.525 $\pm$ 2.751 (12.131, 23.210) \\
\multicolumn{1}{c}{CPM (ec)             }  & \multicolumn{1}{c}{ 10.537 $\pm$ 2.131 (9.595, 18.159) } & 13.456 $\pm$ 2.393 (12.366, 21.976) \\
\multicolumn{1}{c}{CPM (cc)             }  & \multicolumn{1}{c}{ \underline{10.074 $\pm$ 1.307} (\textcolor{blue}{\textbf{9.370}}, 14.741) } & \underline{12.931 $\pm$ 1.669} (\textbf{12.065}, 18.895) \\
\multicolumn{1}{c}{CPM (cnu)            }  & \multicolumn{1}{c}{ 10.895 $\pm$ 4.402 (\underline{9.463}, 26.749) } & 13.775 $\pm$ 4.633 (12.194, 30.465) \\
\multicolumn{1}{c}{CPM (cns)            }  & \multicolumn{1}{c}{ 11.262 $\pm$ 4.585 (9.623, 27.771) } & 14.156 $\pm$ 4.822 (12.406, 31.521) \\
\hline
\multicolumn{1}{c}{PNA-S              }  & \multicolumn{1}{c}{17.217                                                            } & 21.008 \\
\multicolumn{1}{c}{PNA-S (a)              }  & \multicolumn{1}{c}{\underline{12.267 $\pm$ 0.948} (11.002, 14.480) } & 15.338 $\pm$ 1.079 (13.914, 18.110)   \\
\multicolumn{1}{c}{PNA-S (g)              }  & \multicolumn{1}{c}{12.305 $\pm$ 1.246 (10.814, 15.267)} & 15.474 $\pm$ 1.357 (13.897, 18.777)  \\
\multicolumn{1}{c}{PNA-S (tm)              }  & \multicolumn{1}{c}{\textbf{11.537 $\pm$ 0.727} (\textbf{10.639}, 12.890) } & \textbf{14.466 $\pm$ 0.814} (\textbf{13.322}, 15.853)  \\
\multicolumn{1}{c}{PNA-S (dc)              }  & \multicolumn{1}{c}{12.211 $\pm$ 1.119 (10.751, 14.012)  } & 15.389 $\pm$ 1.289 (13.619, 17.191)    \\
\multicolumn{1}{c}{PNA-S (ec)              }  & \multicolumn{1}{c}{12.124 $\pm$ 1.307 (10.956, 16.118) } & \underline{15.270 $\pm$ 1.466} (13.865, 19.509)   \\
\multicolumn{1}{c}{PNA-S (cc)              }  & \multicolumn{1}{c}{12.166 $\pm$ 0.898 (\underline{10.727}, 13.895)  } & 15.413 $\pm$ 1.098 (\underline{13.615}, 17.309)  \\
\multicolumn{1}{c}{PNA-S (cnu)              }  & \multicolumn{1}{c}{ 12.608 $\pm$ 1.204 (10.935, 15.944)  } &  15.799 $\pm$ 1.378 (13.827, 19.684)   \\
\multicolumn{1}{c}{PNA-S (cns)              }  & \multicolumn{1}{c}{ 12.509 $\pm$ 0.963 (11.152, 14.747)  } & 15.737 $\pm$ 1.172 (13.960, 18.528)  \\
\hline
\multicolumn{1}{c}{PNA-V             }   & \multicolumn{1}{c}{20.109                                                    } & 25.113\\
\multicolumn{1}{c}{PNA-V (a)              }  & \multicolumn{1}{c}{15.979 $\pm$ 3.730 (11.607, 26.483)  } & 19.617 $\pm$ 4.111 (14.575, 30.983)    \\
\multicolumn{1}{c}{PNA-V (g)              }  & \multicolumn{1}{c}{\underline{14.570 $\pm$ 4.115} (\textbf{11.014}, 26.962) } & \underline{18.020 $\pm$ 4.329} (\textbf{13.984}, 30.811)   \\
\multicolumn{1}{c}{PNA-V (tm)              }  & \multicolumn{1}{c}{15.450 $\pm$ 4.313 (\underline{11.259}, 26.906)  } & 19.030 $\pm$ 4.804 (\underline{14.273}, 32.059) \\
\multicolumn{1}{c}{PNA-V (dc)              }  & \multicolumn{1}{c}{\textbf{14.300 $\pm$ 2.111} (11.475, 17.750)  } &  \textbf{17.900 $\pm$ 2.588} (14.313, 22.516)   \\
\multicolumn{1}{c}{PNA-V (ec)              }  & \multicolumn{1}{c}{15.385 $\pm$ 3.141 (11.504, 21.710)  } & 19.294 $\pm$ 3.905 (14.779, 27.041)  \\
\multicolumn{1}{c}{PNA-V (cc)              }  & \multicolumn{1}{c}{ 15.288 $\pm$ 3.023 (11.709, 20.592)  } & 18.845 $\pm$ 3.378 (14.770, 24.803)  \\
\multicolumn{1}{c}{PNA-V (cnu)              }  & \multicolumn{1}{c}{  16.800 $\pm$ 5.053 (12.352, 28.022)  } &  20.470 $\pm$ 5.736 (15.180, 32.794)    \\
\multicolumn{1}{c}{PNA-V (cns)              }  & \multicolumn{1}{c}{ 15.910 $\pm$ 5.976 (12.507, 36.704)  } & 19.671 $\pm$ 6.875 (15.478, 43.397)  \\
\hline
\multicolumn{1}{c}{RegGNN                }   & \multicolumn{1}{c}{{9.768}                                                    } & \textcolor{blue}{\textbf{12.270}}\\
\multicolumn{1}{c}{RegGNN (a)  }   & \multicolumn{1}{c}{10.360 $\pm$ 1.090 (9.624, 14.027) } & 12.997 $\pm$ 1.278 (12.158, 17.335)  \\
\multicolumn{1}{c}{RegGNN (g)}   & \multicolumn{1}{c}{10.032 $\pm$ 0.330 (9.576, 10.790) } & 12.563 $\pm$ 0.310 (12.148, 13.311)  \\
\multicolumn{1}{c}{RegGNN (tm)}   & \multicolumn{1}{c}{9.820 $\pm$ 0.512 (9.525, 11.613)  } & \underline{12.378 $\pm$ 0.528} (12.116, 14.259) \\
\multicolumn{1}{c}{RegGNN (dc)}   & \multicolumn{1}{c}{{{\textbf{9.714 $\pm$ 0.332}} (9.485, 10.634)} } & 12.531 $\pm$ 0.458 (\textcolor{blue}{\textbf{12.064}}, 13.716)  \\
\multicolumn{1}{c}{RegGNN (ec) }   & \multicolumn{1}{c}{9.815 $\pm$ 0.245 (9.469, 10.334) } & 12.609 $\pm$ 0.334 (12.171, 13.286)  \\
\multicolumn{1}{c}{RegGNN (cc) }   & \multicolumn{1}{c}{9.777 $\pm$ 0.391 (9.461, 10.757)  } & 12.516 $\pm$ 0.425 (12.121, 13.523)  \\ 
\multicolumn{1}{c}{RegGNN (cnu) }   & \multicolumn{1}{c}{9.745 $\pm$ 0.451 ({\textbf{9.438}}, 11.018) } & 12.482 $\pm$ 0.473 (\underline{12.085}, 13.758)   \\
\multicolumn{1}{c}{RegGNN (cns) }   & \multicolumn{1}{c}{\underline{9.716 $\pm$ 0.292} (\underline{9.452}, 10.474) } & 12.466 $\pm$ 0.217 (12.207, 12.982) \\ 
\hline
\multicolumn{3}{c}{\cellcolor[HTML]{cFFFa9}  \textbf{NT (VIQ)}}\\ \hline
\multicolumn{1}{c}{CPM                    }  & \multicolumn{1}{c}{\textbf{9.517}} & \textbf{12.049}\\
\multicolumn{1}{c}{CPM (a)              }  & \multicolumn{1}{c}{ 11.035 $\pm$ 4.970 (9.481, 28.895) } & 13.728 $\pm$ 5.243 (12.031, 32.518) \\
\multicolumn{1}{c}{CPM (g)              }  & \multicolumn{1}{c}{ 10.363 $\pm$ 1.682 (9.549, 16.146) } & 13.253 $\pm$ 2.126 (12.236, 20.633) \\
\multicolumn{1}{c}{CPM (tm)     }  & \multicolumn{1}{c}{ 10.194 $\pm$ 2.316 (9.503, 18.543) } & 12.922 $\pm$ 2.928 (12.035, 23.477) \\
\multicolumn{1}{c}{CPM (dc)             }  & \multicolumn{1}{c}{ 10.065 $\pm$ 1.760 (\underline{9.391}, 16.386)  } & 12.767 $\pm$ 2.386 (\underline{11.877}, 21.332) 
 \\
\multicolumn{1}{c}{CPM (ec)             }  & \multicolumn{1}{c}{ \underline{9.942 $\pm$ 1.768} (\textcolor{blue}{\textbf{9.323}}, 16.275)} & \underline{12.614 $\pm$ 2.209} (\textcolor{blue}{\textbf{11.828}}, 20.532)  \\
\multicolumn{1}{c}{CPM (cc)             }  & \multicolumn{1}{c}{ 10.233 $\pm$ 1.264 (9.514, 14.667) } & 12.911 $\pm$ 1.595 (12.035, 18.505) \\
\multicolumn{1}{c}{CPM (cnu)            }  & \multicolumn{1}{c}{ 10.544 $\pm$ 3.585 (9.480, 23.469) } & 13.256 $\pm$ 4.058 (12.022, 27.885) \\
\multicolumn{1}{c}{CPM (cns)            }  & \multicolumn{1}{c}{ 10.583 $\pm$ 3.560 (9.526, 23.416) } & 13.326 $\pm$ 3.891 (12.108, 27.343) \\
\hline
\multicolumn{1}{c}{PNA-S              }  & \multicolumn{1}{c}{12.838                                                      } & 16.130\\
\multicolumn{1}{c}{PNA-S (a)              }  & \multicolumn{1}{c}{\underline{11.846 $\pm$ 0.942} (10.795, 13.764) } & \underline{15.057 $\pm$ 1.099} (13.824, 17.279)   \\
\multicolumn{1}{c}{PNA-S (g)              }  & \multicolumn{1}{c}{12.439 $\pm$ 1.183 (10.913, 14.774) } & 15.648 $\pm$ 1.322 (13.895, 18.469)  \\
\multicolumn{1}{c}{PNA-S (tm)              }  & \multicolumn{1}{c}{12.489 $\pm$ 3.099 (10.759, 23.367) } & 15.884 $\pm$ 4.323 (13.983, 31.268)  \\
\multicolumn{1}{c}{PNA-S (dc)              }  & \multicolumn{1}{c}{\textbf{11.694 $\pm$ 0.814} (\textbf{10.302}, 12.901)  } & \textbf{14.707 $\pm$ 0.904} (\textbf{13.126}, 16.115) \\
\multicolumn{1}{c}{PNA-S (ec)              }  & \multicolumn{1}{c}{12.091 $\pm$ 1.006 (\underline{10.543}, 14.032) } & 15.122 $\pm$ 1.061 (\underline{13.436}, 17.200)   \\
\multicolumn{1}{c}{PNA-S (cc)              }  & \multicolumn{1}{c}{13.074 $\pm$ 1.693 (11.258, 17.486) } & 16.326 $\pm$ 1.795 (14.474, 20.701) \\
\multicolumn{1}{c}{PNA-S (cnu)              }  & \multicolumn{1}{c}{ 12.682 $\pm$ 1.308 (10.857, 15.196)  } &  15.950 $\pm$ 1.536 (13.795, 18.775)   \\
\multicolumn{1}{c}{PNA-S (cns)              }  & \multicolumn{1}{c}{ 12.014 $\pm$ 0.859 (10.545, 14.031)  } & 15.141 $\pm$ 0.944 (13.503, 17.302)  \\
\hline
\multicolumn{1}{c}{PNA-V              }  & \multicolumn{1}{c}{14.695                                               } & 18.903\\
\multicolumn{1}{c}{PNA-V (a)              }  & \multicolumn{1}{c}{\textbf{14.107 $\pm$ 2.081} (11.923, 19.482)  } & \textbf{17.696 $\pm$ 2.693} (14.832, 25.550)  \\
\multicolumn{1}{c}{PNA-V (g)              }  & \multicolumn{1}{c}{14.983 $\pm$ 5.211 (11.639, 32.479)  } & 18.681 $\pm$ 6.224 (14.715, 39.771)  \\
\multicolumn{1}{c}{PNA-V (tm)              }  & \multicolumn{1}{c}{15.489 $\pm$ 4.211 (11.717, 27.980)  } & 19.157 $\pm$ 4.775 (14.624, 33.183)   \\
\multicolumn{1}{c}{PNA-V (dc)              }  & \multicolumn{1}{c}{\underline{14.332 $\pm$ 2.931} (\textbf{11.188}, 20.545)  } & \underline{18.170 $\pm$ 3.917} (\textbf{14.284}, 26.511)   \\
\multicolumn{1}{c}{PNA-V (ec)              }  & \multicolumn{1}{c}{14.924 $\pm$ 3.424 (11.360, 21.703)  } & 18.539 $\pm$ 4.185 (\underline{14.392}, 27.289)  \\
\multicolumn{1}{c}{PNA-V (cc)              }  & \multicolumn{1}{c}{ 16.049 $\pm$ 4.185 (\underline{11.237}, 24.616)  } & 20.035 $\pm$ 4.966 (14.408, 29.373)   \\
\multicolumn{1}{c}{PNA-V (cnu)              }  & \multicolumn{1}{c}{ 14.805 $\pm$ 2.587 (11.890, 21.107)  } & 18.344 $\pm$ 2.883 (15.171, 25.473)  \\
\multicolumn{1}{c}{PNA-V (cns)              }  & \multicolumn{1}{c}{14.915 $\pm$ 2.845 (11.898, 22.045)  } & 18.674 $\pm$ 3.612 (14.895, 28.544) \\
\hline
\multicolumn{1}{c}{RegGNN                 }  & \multicolumn{1}{c}{10.195                                                      } & 13.044\\
\multicolumn{1}{c}{RegGNN  (a)   }  & \multicolumn{1}{c}{9.587 $\pm$ 0.206 (9.477, 10.311)  } &12.223 $\pm$ 0.299 (12.054, 13.261)  \\
\multicolumn{1}{c}{RegGNN  (g) }  & \multicolumn{1}{c}{9.779 $\pm$ 0.126 (9.551, 9.964) } & 12.530 $\pm$ 0.199 (12.190, 12.803)  \\
\multicolumn{1}{c}{RegGNN  (tm) }  & \multicolumn{1}{c}{{\textcolor{blue}{\textbf{9.514 $\pm$ 0.036}} (\underline{9.471}, 9.594) }} & \textcolor{blue}{\textbf{12.041 $\pm$ 0.035}} ({\textbf{12.004}}, 12.140) \\
\multicolumn{1}{c}{RegGNN  (dc) }  & \multicolumn{1}{c}{9.639 $\pm$ 0.161 (9.468, 10.000)  } & 12.213 $\pm$ 0.235 (\underline{12.008}, 12.707) \\
\multicolumn{1}{c}{RegGNN  (ec)  }  & \multicolumn{1}{c}{9.599 $\pm$ 0.243 ({\textbf{9.453}}, 10.189) } & 12.199 $\pm$ 0.274 (12.020, 12.846) \\
\multicolumn{1}{c}{RegGNN (cc) }   & \multicolumn{1}{c}{9.711 $\pm$ 0.193 (9.499, 10.161) } &12.313 $\pm$ 0.284 (12.023, 13.062) \\ 
\multicolumn{1}{c}{RegGNN (cnu) }   & \multicolumn{1}{c}{{\underline{9.521 $\pm$ 0.042} (9.473, 9.651)}  } &  \underline{12.134 $\pm$ 0.050} (12.050, 12.266)  \\
\multicolumn{1}{c}{RegGNN (cns) }   & \multicolumn{1}{c}{ 9.581 $\pm$ 0.153 (9.497, 10.109) } & 12.236 $\pm$ 0.199 (12.102, 12.907)    \\ 
\hline
\end{tabular}

\caption{Comparison of regression methods on the NT cohort. The best performing method for each architecture is bold while the second best is underlined. The mean $\pm$ standard deviation as well as minima and maxima over $k=2,\dots,15$ (in brackets) are given. The overall best performing method according to mean error and the best sample selection performance are indicated in blue. Abbreviations are: (a) absolute Euclidean distance, (g) geometric Log-Euclidean distance, (tm) full tangent matrix, (dc) degree centrality, (ec) eigenvector centrality, (cc) closeness centrality, (cnu) concatination unscaled, (cns) concatination scaled.} 
\label{tab:results_nt}
\end{table*}

\begin{table*}[ht]
\scriptsize
\centering
\begin{tabular}{ccc}
\hline
\multicolumn{1}{c}{\textbf{Method}}          & \multicolumn{1}{c}{\textbf{Mean MAE $\pm$ std (min, max)}}                     & \textbf{Mean RMSE $\pm$ std (min, max)}\\ \hline
\multicolumn{3}{c}{\cellcolor[HTML]{Fce3e3}  \textbf{ASD (FIQ)}}\\ \hline
\multicolumn{1}{c}{CPM          }  & \multicolumn{1}{c}{ \underline{12.533} } &  15.965 \\
\multicolumn{1}{c}{CPM (a)      }  & \multicolumn{1}{c}{ 13.673 $\pm$ 3.748 (12.082, 26.912) } & 16.783 $\pm$ 4.048 (\underline{15.043}, 31.074) \\
\multicolumn{1}{c}{CPM (g)      }  & \multicolumn{1}{c}{ 13.267 $\pm$ 1.367 (12.166, 17.792) } & 16.728 $\pm$ 1.672 (15.433, 22.297) \\
\multicolumn{1}{c}{CPM (tm)     }  & \multicolumn{1}{c}{ \textbf{12.464 $\pm$ 1.090} (12.093, 16.386) } & \textbf{15.522 $\pm$ 1.336} (15.084, 20.333) \\
\multicolumn{1}{c}{CPM (dc)     }  & \multicolumn{1}{c}{ 12.644 $\pm$ 1.498 (\underline{12.082}, 18.027) } & \underline{15.685 $\pm$ 1.723} (15.116, 21.887) \\
\multicolumn{1}{c}{CPM (ec)     }  & \multicolumn{1}{c}{ 12.825 $\pm$ 1.400 (\textbf{12.074}, 17.527) } & 15.922 $\pm$ 1.654 (\textcolor{blue}{\textbf{15.017}}, 21.252) \\
\multicolumn{1}{c}{CPM (cc)     }  & \multicolumn{1}{c}{ 13.310 $\pm$ 1.331 (12.430, 17.821) } & 16.703 $\pm$ 1.654 (15.618, 22.335) \\
\multicolumn{1}{c}{CPM (cnu)    }  & \multicolumn{1}{c}{ 12.934 $\pm$ 2.460 (12.094, 21.790) } & 16.039 $\pm$ 2.817 (15.086, 26.182) \\
\multicolumn{1}{c}{CPM (cns)    }  & \multicolumn{1}{c}{ 12.742 $\pm$ 1.527 (12.166, 18.129) } & 15.796 $\pm$ 1.838 (15.077, 22.298) \\ \hline
\multicolumn{1}{c}{PNA-S              }  & \multicolumn{1}{c}{17.162                                                            } & 22.023 \\
\multicolumn{1}{c}{PNA-S (a)              }  & \multicolumn{1}{c}{15.260 $\pm$ 0.865 (13.947, 17.049)  } & 18.768 $\pm$ 0.940 (17.527, 20.665) \\
\multicolumn{1}{c}{PNA-S (g)              }  & \multicolumn{1}{c}{15.025 $\pm$ 1.321 (13.207, 17.762)  } & 18.844 $\pm$ 1.641 (16.642, 22.009)   \\
\multicolumn{1}{c}{PNA-S (tm)              }  & \multicolumn{1}{c}{14.781 $\pm$ 1.182 (\textbf{13.168}, 18.488)  } & 18.630 $\pm$ 1.496 (16.522, 23.426)  \\
\multicolumn{1}{c}{PNA-S (dc)              }  & \multicolumn{1}{c}{14.248 $\pm$ 0.533 (13.595, 15.651)   } & 17.715 $\pm$ 0.723 (16.654, 19.644)  \\
\multicolumn{1}{c}{PNA-S (ec)              }  & \multicolumn{1}{c}{14.758 $\pm$ 1.277 (\underline{13.207}, 18.534)  } & 18.263 $\pm$ 1.608 (\textbf{16.440}, 23.134)    \\
\multicolumn{1}{c}{PNA-S (cc)              }  & \multicolumn{1}{c}{14.558 $\pm$ 0.666 (13.824, 16.586)  } & 18.126 $\pm$ 0.894 (17.296, 20.806)  \\
\multicolumn{1}{c}{PNA-S (cnu)              }  & \multicolumn{1}{c}{ \underline{14.038 $\pm$ 0.434} (13.353, 14.915)   } &  \textbf{17.438 $\pm$ 0.580} (\underline{16.469}, 18.642)    \\
\multicolumn{1}{c}{PNA-S (cns)              }  & \multicolumn{1}{c}{ \textbf{14.023 $\pm$ 0.401} (13.465, 14.782)   } & \underline{17.620 $\pm$ 0.550} (16.853, 18.605)  \\ \hline

\multicolumn{1}{c}{PNA-V              }  & \multicolumn{1}{c}{\underline{15.671}                                                            } & 20.085 \\

\multicolumn{1}{c}{PNA-V (a)              }  & \multicolumn{1}{c}{17.386 $\pm$ 2.428 (14.453, 23.263)  } & 21.234 $\pm$ 2.922 (18.006, 28.505) \\
\multicolumn{1}{c}{PNA-V (g)              }  & \multicolumn{1}{c}{17.611 $\pm$ 2.228 (14.770, 23.572) } & 22.035 $\pm$ 3.023 (18.422, 30.646)    \\
\multicolumn{1}{c}{PNA-V (tm)              }  & \multicolumn{1}{c}{18.458 $\pm$ 5.559 (\textbf{13.770}, 32.776)   } & 23.112 $\pm$ 6.498 (\underline{17.326}, 38.914)  \\
\multicolumn{1}{c}{PNA-V (dc)              }  & \multicolumn{1}{c}{16.358 $\pm$ 2.793 (14.045, 24.543)   } & 20.325 $\pm$ 3.344 (17.355, 29.689)   \\
\multicolumn{1}{c}{PNA-V (ec)              }  & \multicolumn{1}{c}{17.085 $\pm$ 4.124 (\underline{13.778}, 28.115)  } & 21.095 $\pm$ 4.885 (\textbf{17.101}, 33.584)   \\
\multicolumn{1}{c}{PNA-V (cc)              }  & \multicolumn{1}{c}{16.172 $\pm$ 1.181 (14.541, 18.242)  } & \underline{19.942 $\pm$ 1.346} (17.970, 22.102)  \\
\multicolumn{1}{c}{PNA-V (cnu)              }  & \multicolumn{1}{c}{ 19.552 $\pm$ 6.530 (13.840, 33.875)   } & 24.470 $\pm$ 8.514 (17.399, 44.199)  \\
\multicolumn{1}{c}{PNA-V (cns)              }  & \multicolumn{1}{c}{ \textbf{15.601 $\pm$ 2.041} (13.820, 20.748)    } & \textbf{19.406 $\pm$ 2.359} (17.484, 25.427)  \\ \hline

\multicolumn{1}{c}{RegGNN                 }  & \multicolumn{1}{c}{12.564 } & 15.624 \\
\multicolumn{1}{c}{RegGNN (a)   }  & \multicolumn{1}{c}{12.588 $\pm$ 0.541 (12.170, 13.825)   }          & 15.628 $\pm$ 0.598 (15.102, 17.006) \\ 
\multicolumn{1}{c}{RegGNN (g) }  & \multicolumn{1}{c}{12.977 $\pm$ 1.073 (12.137, 16.292)   } & 16.194 $\pm$ 1.358 (15.146, 20.300)  \\
\multicolumn{1}{c}{RegGNN (tm) }  & \multicolumn{1}{c}{\textcolor{blue}{\textbf{12.148 $\pm$ 0.072}} (12.078, 12.379)  } & \textcolor{blue}{\textbf{15.130 $\pm$ 0.067}} (15.074, 15.355)  \\
\multicolumn{1}{c}{RegGNN (dc) }  & \multicolumn{1}{c}{\underline{12.247 $\pm$ 0.167} (\underline{12.073}, 12.667)}  & \underline{15.214 $\pm$ 0.178} (\underline{15.058}, 15.693)  \\
\multicolumn{1}{c}{RegGNN (ec)  }  & \multicolumn{1}{c}{12.361 $\pm$ 0.246 (12.106, 12.905)  }          & 15.320 $\pm$ 0.265 ({\textbf{15.040}, 15.895)} \\ 
\multicolumn{1}{c}{RegGNN (cc) }   & \multicolumn{1}{c}{12.720 $\pm$ 0.466 (12.134, 14.008)  } & 15.756 $\pm$ 0.560 (15.074, 17.321)    \\ 
\multicolumn{1}{c}{RegGNN (cnu) }   & \multicolumn{1}{c}{12.301 $\pm$ 0.226 (\textcolor{blue}{\textbf{12.064}}, 12.804)  } & 15.331 $\pm$ 0.250 (15.120, 15.881)    \\
\multicolumn{1}{c}{RegGNN (cns) }   & \multicolumn{1}{c}{ 12.300 $\pm$ 0.159 (12.132, 12.763) } &15.306 $\pm$ 0.201 (15.114, 15.894)   \\ 
\hline
\multicolumn{3}{c}{\cellcolor[HTML]{F6a1a2} \textbf{ASD (VIQ)}}\\ \hline

\multicolumn{1}{c}{CPM          }  & \multicolumn{1}{c}{ \underline{14.171} } &  18.834 \\
\multicolumn{1}{c}{CPM (a)      }  & \multicolumn{1}{c}{ 15.006 $\pm$ 5.390 (12.920, 34.315) } & 19.034 $\pm$ 5.458 (16.887, 38.586) \\
\multicolumn{1}{c}{CPM (g)      }  & \multicolumn{1}{c}{ 14.336 $\pm$ 3.490 (\textbf{12.669}, 26.704) } & \underline{18.646 $\pm$ 3.660} (\underline{16.845}, 31.560) \\
\multicolumn{1}{c}{CPM (tm)     }  & \multicolumn{1}{c}{ 15.496 $\pm$ 2.494 (13.694, 21.575) } & 20.191 $\pm$ 3.140 (17.994, 28.150) \\
\multicolumn{1}{c}{CPM (dc)     }  & \multicolumn{1}{c}{ 14.449 $\pm$ 1.882 (13.289, 19.795) } & 18.767 $\pm$ 2.166 (17.421, 24.648) \\
\multicolumn{1}{c}{CPM (ec)     }  & \multicolumn{1}{c}{ \textbf{13.697 $\pm$ 1.389} (\underline{12.824}, 18.258) } & \textbf{17.985 $\pm$ 1.737} (\textbf{16.838}, 23.693) \\
\multicolumn{1}{c}{CPM (cc)     }  & \multicolumn{1}{c}{ 14.954 $\pm$ 3.088 (13.432, 25.694) } & 19.613 $\pm$ 3.506 (17.508, 31.570) \\
\multicolumn{1}{c}{CPM (cnu)    }  & \multicolumn{1}{c}{ 14.698 $\pm$ 1.565 (13.046, 19.980) } & 18.917 $\pm$ 1.715 (17.412, 24.674) \\
\multicolumn{1}{c}{CPM (cns)    }  & \multicolumn{1}{c}{ 15.417 $\pm$ 2.761 (13.593, 22.869) } & 20.041 $\pm$ 4.128 (17.648, 32.850) \\ \hline

\multicolumn{1}{c}{PNA-S             }   & \multicolumn{1}{c}{19.955                                            } & 25.848\\
\multicolumn{1}{c}{PNA-S (a)              }  & \multicolumn{1}{c}{15.993 $\pm$ 1.096 (\textbf{14.125}, 18.861)   } & 20.500 $\pm$ 1.284 (\underline{18.349}, 23.726) \\
\multicolumn{1}{c}{PNA-S (g)              }  & \multicolumn{1}{c}{\textbf{15.540 $\pm$ 1.176} (14.139, 18.017) } & \textbf{20.037 $\pm$ 1.548} (\textbf{18.256}, 23.559)   \\
\multicolumn{1}{c}{PNA-S (tm)              }  & \multicolumn{1}{c}{16.766 $\pm$ 1.330 (\underline{15.020}, 20.365)   } & 21.548 $\pm$ 1.583 (19.513, 25.735)   \\
\multicolumn{1}{c}{PNA-S (dc)              }  & \multicolumn{1}{c}{16.820 $\pm$ 2.161 (14.332, 21.360)    } & 21.386 $\pm$ 2.303 (18.685, 25.724)  \\
\multicolumn{1}{c}{PNA-S (ec)              }  & \multicolumn{1}{c}{16.445 $\pm$ 1.598 (14.518, 21.308)   } & 21.150 $\pm$ 1.900 (18.760, 26.560)    \\
\multicolumn{1}{c}{PNA-S (cc)              }  & \multicolumn{1}{c}{16.113 $\pm$ 1.458 (14.600, 19.985)   } & 21.012 $\pm$ 1.715 (19.057, 25.231)   \\
\multicolumn{1}{c}{PNA-S (cnu)              }  & \multicolumn{1}{c}{ 16.359 $\pm$ 1.450 (14.661, 20.737)    } &  20.880 $\pm$ 1.714 (18.905, 26.173)    \\
\multicolumn{1}{c}{PNA-S (cns)              }  & \multicolumn{1}{c}{ \underline{15.887 $\pm$ 0.773} (14.621, 17.091)    } & \underline{20.282 $\pm$ 0.843} (18.785, 21.847)  \\ \hline

\multicolumn{1}{c}{PNA-V             }   & \multicolumn{1}{c}{22.518                                            } & {28.804}\\
\multicolumn{1}{c}{PNA-V (a)              }  & \multicolumn{1}{c}{ 18.145 $\pm$ 3.017 (14.837, 26.420)  } & 22.937 $\pm$ 3.373 (19.233, 31.651)  \\
\multicolumn{1}{c}{PNA-V (g)              }  & \multicolumn{1}{c}{ \textbf{16.906 $\pm$ 1.566} (14.975, 21.123)  } & \textbf{21.791 $\pm$ 2.210} (19.658, 28.209)  \\
\multicolumn{1}{c}{PNA-V (tm)              }  & \multicolumn{1}{c}{ 19.040 $\pm$ 3.597 (15.631, 30.284)  } & 24.162 $\pm$ 4.053 (19.972, 36.701)  \\
\multicolumn{1}{c}{PNA-V (dc)              }  & \multicolumn{1}{c}{ 19.734 $\pm$ 4.936 (15.055, 33.387)   } & 25.088 $\pm$ 6.380 (19.898, 43.769)   \\
\multicolumn{1}{c}{PNA-V (ec)              }  & \multicolumn{1}{c}{ 17.837 $\pm$ 2.000 (14.973, 21.401)  } &  22.721 $\pm$ 2.341 (19.373, 27.477)  \\
\multicolumn{1}{c}{PNA-V (cc)              }  & \multicolumn{1}{c}{ 18.663 $\pm$ 4.268 (14.969, 32.436)  } & 24.016 $\pm$ 5.063 (19.679, 40.173)  \\
\multicolumn{1}{c}{PNA-V (cnu)              }  & \multicolumn{1}{c}{  \underline{17.314 $\pm$ 2.329} (14.660, 22.610)  } & \underline{22.128 $\pm$ 2.772} (\underline{19.026}, 28.786)  \\
\multicolumn{1}{c}{PNA-V (cns)              }  & \multicolumn{1}{c}{ 17.714 $\pm$ 2.833 (14.697, 24.781)   } & 22.630 $\pm$ 3.748 (\textbf{18.958}, 32.707)  \\ \hline

\multicolumn{1}{c}{RegGNN                }   & \multicolumn{1}{c}{\textcolor{blue}{\textbf{13.090}}                                } & \textcolor{blue}{\textbf{17.250}}\\
\multicolumn{1}{c}{RegGNN (a)}   & \multicolumn{1}{c}{{13.356 $\pm$ 0.443 (12.834, 14.402)    }} & \underline{17.272 $\pm$ 0.480} (\textcolor{blue}{\textbf{16.782}}, 18.502)  \\
\multicolumn{1}{c}{RegGNN (g)}   & \multicolumn{1}{c}{{\underline{13.316 $\pm$ 0.545} (\textcolor{blue}{\textbf{12.687}}, 14.700)  }} & 17.474 $\pm$ 0.663 (16.877, 19.186)  \\
\multicolumn{1}{c}{RegGNN (tm)}   & \multicolumn{1}{c}{14.110 $\pm$ 1.237 (13.194, 17.014)   } &18.288 $\pm$ 1.656 (17.130, 22.128)  \\
\multicolumn{1}{c}{RegGNN (dc)}   & \multicolumn{1}{c}{14.516 $\pm$ 2.043 (13.369, 19.931)                     } &18.381 $\pm$ 2.261 (17.140, 24.445)  \\
\multicolumn{1}{c}{RegGNN (ec)}   & \multicolumn{1}{c}{13.419 $\pm$ 0.705 (12.766, 15.061)      } & 17.526 $\pm$ 0.945 (\underline{16.815}, 19.958)   \\ 
\multicolumn{1}{c}{RegGNN (cc) }   & \multicolumn{1}{c}{14.021 $\pm$ 1.001 (13.149, 16.812)  } &  18.731 $\pm$ 1.242 (17.563, 21.916)   \\ 
\multicolumn{1}{c}{RegGNN (cnu) }   & \multicolumn{1}{c}{14.001 $\pm$ 0.605 (\underline{12.762}, 14.885)  } & 18.084 $\pm$ 0.569 (17.022, 18.924)   \\
\multicolumn{1}{c}{RegGNN (cns) }   & \multicolumn{1}{c}{ 14.020 $\pm$ 0.440 (12.950, 14.650)  } & 18.007 $\pm$ 0.434 (17.008, 18.571)   \\ 
 \hline
\end{tabular}
\caption{Comparison of regression methods on the ASD cohort. The best performing method for each architecture is bold while the second best is underlined. The mean $\pm$ standard deviation as well as minima and maxima over $k=2,\dots,15$ (in brackets) are given. The overall best performing method according to mean error and the best sample selection performance are indicated in blue. Abbreviations are: (a) absolute Euclidean distance, (g) geometric Log-Euclidean distance, (tm) full tangent matrix, (dc) degree centrality, (ec) eigenvector centrality, (cc) closeness centrality, (cnu) concatination unscaled, (cns) concatination scaled.}
\label{tab:results_asd}
\end{table*}

\textbf{Dataset.}
    We used samples from the Autism Brain Imaging Data Exchange (ABIDE) Preprocessed dataset \citep{Craddock:2013} for our experiments. It contains data from 16 imaging sites, preprocessed by five different teams using four pipelines: the Connectome Computation System (CCS), the Configurable Pipeline for the Analysis of Connectomes (CPAC), the Data Processing Assistant for rs-fMRI (DPARSF) and the NeuroImaging Analysis Kit. The preprocessed data sets are available online\footnote{\href{http://preprocessed-connectomes-project.org/abide/}{http://preprocessed-connectomes-project.org/abide/}}. To account for possible biases due to differences in sites, we used randomly sampled subsets of the available data for both cohorts; the same sets were also used by \citep{Dryburgh:2020}. The NT cohort consisted of 226 subjects (with mean age = (15 $\pm$ 3.6)), while the ASD cohort was made up of 202 subjects (with mean age = (15.4 $\pm$ 3.8)). FIQ and VIQ scores in the NT cohort have means 111.573 $\pm$ 12.056 and 112.787 $\pm$ 12.018, whereas FIQ and VIQ scores in the ASD cohort have means 106.102 $\pm$ 15.045 and 103.005 $\pm$ 16.874, respectively. The brain connectomes were obtained from resting-state functional magnetic resonance imaging using the parcellation from \citep{TzourioMazoyer:2002} with 116 ROIs. The functional connectomes are represented by 116-by-116 matrices, whose entry in row $i$ and column $j$ is the Pearson correlation between the average rs-fMRI signal measured in ROI $i$ and ROI $j$.
    
\textbf{Software.}
    All experiments are done in Python 3.7.10. We used Scikit-learn 0.24.2~\citep{Pedregosa:2011} for machine learning models and PyTorch Geometric 1.6.3~\citep{Fey:2019} for graph neural network implementations.
    For Riemannian geometric computations in the SPD space, we used the SPD class from the Morphomatics package of \citep{Morphomatics}. To extract the graph topological features from the tangent matrices we used NetworkX~\citep{Hagberg:2008}.

\textbf{Parameter settings.}
    We trained our method with Adam optimizer~\citep{Kingma:2017} for 100 epochs with a learning rate of 0.001 and weight decay at 0.0005 based on our empirical observations. Furthermore, we set $d_1 := 64$ in RegGNN. The dropout rate after the first graph convolutional layer was set to 0.1.
    To regularize the adjacency matrices, we used $\mu = 10^{-10}$ in (\ref{eq:regularize}). 
    In order to explore the parameter space for the number of selected training samples $k$, we varied it between $2$ and $15$.

\textbf{Evaluation and comparison methods.}
    To test the generalizability and robustness of our method, we used 3-fold cross-validation on both NT and ASD cohorts separately for both FIQ and VIQ prediction. We report the mean absolute error (MAE) and the root mean squared error (RMSE) for all methods. For the sample selection methods, we additionally give the mean, standard deviation, minima and maxima over all tested $k=2,\dots,15$ to test our sample selection methods sensitivity to the selection of $k$.
    
    To benchmark against our method, we chose state-of-the-art methods from both deep learning and machine learning. The first baseline was CPM~\citep{Shen:2017}, which was specifically designed for behavioral score prediction on brain connectomes; the second being PNA~\citep{Corso:2020}, which outperformed common GNNs on both artificial and real-world benchmark regression tasks (but has not been applied to brain connectomes yet).
    PNA comes with \textit{principal neighborhood aggregation} layers that are defined similarly to graph convolution operations. They are designed to increase the amount of information that is used from the local neighborhoods in the graphs. In our experiments, we inserted PNA layers in our RegGNN architecture. We implemented both a simpler setup with sum aggregation and identity scaling only (denoted by PNA-S), as well as various aggregation (sum, mean, var and max) and scaling (identity, amplification and attenuation) methods (denoted by PNA-V) as detailed in the paper of~\citep{Corso:2020}. The code of both CPM\footnote{\href{https://github.com/esfinn/cpm\_tutorial}{https://github.com/esfinn/cpm\_tutorial}} and PNA\footnote{\href{https://github.com/lukecavabarrett/pna/}{https://github.com/lukecavabarrett/pna/}} is available online. 
    
    In order to assess the effect of the sample selection method, we also always trained each architecture on \emph{all} samples as a baseline. 

\textbf{Evaluation of the sample selection.}
    For each architecture, we compared several methods that can be used as measure of difference in the sample selection (viz., Sec.~\ref{Sec:sample_selection} part $(ii)$) to train the linear mapping $f$.
    
    The first class of methods was the proposed one: we encoded the differences via tangent matrices in the SPD space.
    To identify a good choice for handling the information that is contained in the tangent matrices, we compared several methods. As one option, we trained $f$ on the vectorized upper triangular part (including the diagonal) of the tangent matrix; this method is denoted by (tm). Since the matrices are symmetric, ignoring the lower part speeds up computations while not losing information.
    Further, we used
    degree centrality,
    eigenvector centrality, and closeness centrality (see App.~\ref{Sec:AppA}), and applied them to the tangent matrices; they are denoted by (dc), (ec), and (cc), respectively. 
    \majorrev{Note that during the process, the topology of each connectome is not altered.}
    The mapping $f$ was then trained on the resulting centrality vectors. 
    Additionally, we tested whether the concatenation of the above centrality measures into a single vector is even more informative. To this end, we used both an unscaled and a scaled version, denoted by (cnu) and (cns) respectively. The unscaled version was generated by simple concatenation of the three feature vectors. However, as the three centrality measures have different ranges, we additionally tested scaling each feature vector first. For this, we used min-max scaling. Remember that min-max scaling of a vector $v$ is defined element-wise by $$\widetilde{v_{i}} := \frac{v_{i} - \max(v)}{\max(v) - \min(v)}.$$
    Each centrality vector was scaled before concatenating, which then gave a vector with elements in $[0,1]$ as data for the regression.
    
    We complemented these methods with two baselines.
    In order to check whether the additional directional information that the tangent matrices contain helps, we also tested whether it suffices to train $f$ on the Riemannian geometric distances $d(\mathbf{P}^{s}_i,\mathbf{P}^{s}_j)$ between the connectomes from the train-in group alone; this method is denoted by (g). To assess whether we improve by using the manifold structure of the SPD space at all, we trained $f$ on the Euclidean absolute distance between the upper triangular parts $\widehat{\mathbf{P}}^{s}_i, \widehat{\mathbf{P}}^{s}_j$ of each pair of connectomes $\mathbf{P}^{s}_i, \mathbf{P}^{s}_j$, i.e., on the scalars $\|\widehat{\mathbf{P}}^{s}_i - \widehat{\mathbf{P}}^{s}_j\|_{\textnormal{F}}$ (F standing for the Frobenius norm); we denote this method by (a).
    
    We report the $p$-value between the best performing sample selection method MAE and the baseline MAE according to a t-test for all architectures.

\section{Results and Discussion}
    \begin{figure*}[ht] 
        \centering
        \includegraphics[width=\textwidth]{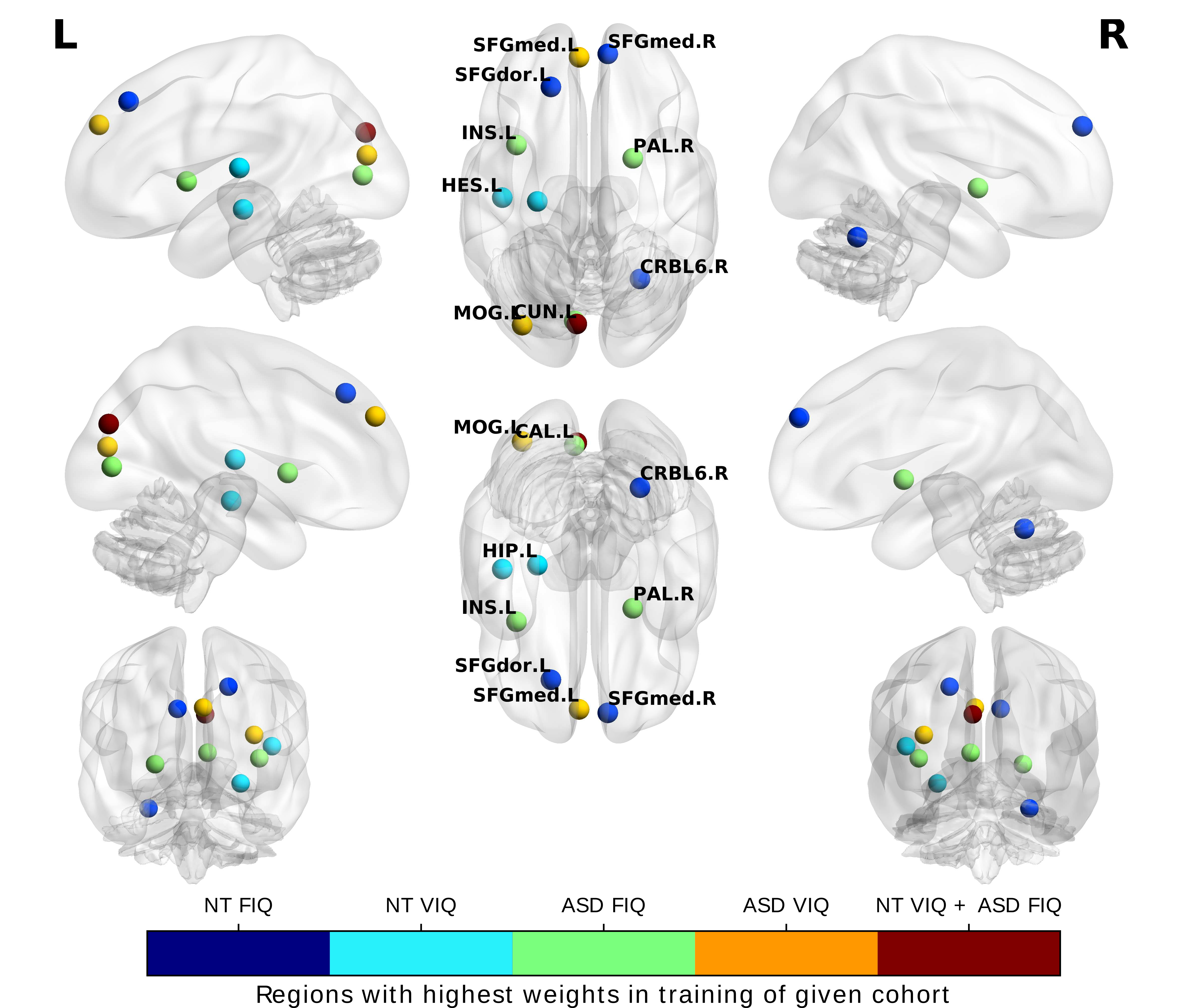}
        \caption{The top three most relevant brain regions for each task according to the learned weights extracted from the last layer of the trained RegGNN. The colors indicate the task. NT: neurotypical subjects. ASD: autism spectrum disorder subjects. FIQ: fluid intelligence quotient. VIQ: verbal intelligence quotient.}
        \label{fig:brain_regions}
    \end{figure*}
    The results for the NT and ASD cohorts are shown in Tables~\ref{tab:results_nt} and \ref{tab:results_asd}, respectively. We observe that while the state-of-the-art machine learning model CPM surpasses naive applications of GNNs in the form of PNA, our RegGNN, paired with sample selection, outperforms CPM in all tasks according to both MAE and RMSE with the exception of the NT (FIQ) task. Improvements by our method are especially visible in the ASD cohort. Interestingly, we see that the results of all methods are worse on the ASD cohort compared to the NT cohort. This was also observed by \citep{Dryburgh:2020}. We hypothesize that the difficulty of predicting IQ scores in ASD cohort might be caused by the inter-subject heterogeneity that is characteristic for ASD~\citep{Tordjman:2018}. Another factor may be that ASD samples from ABIDE are biased towards high-functioning individuals~\citep{Craddock:2013}.

    We observe further that sample selection improved the results for RegGNN in all tasks except ASD (VIQ), and for CPM in the ASD cohort. For PNA based architectures, there are drastic improvements in NT (FIQ) and ASD (VIQ) and incremental improvements in the remaining other tasks. An exception to this is the PNA-V setup for ASD (FIQ), where most models with sample selection perform worse than the one trained on all samples. This might be partly explained by the more complicated structure of PNA with various aggregation models, which might demand more samples for correct training. 
    
    For all models, we see that the minimum MAE over $k$ is lower than the MAE version that was trained on the full data sets even when the mean MAE across $k$ is higher in all tasks. This indicates that improvements are highly likely with fine-tuning of parameter $k$. 
    
    Our experiments did not reveal a clear trend for the value of $k$ for which the minimum was attained. \majorrev{Nevertheless, our observations show that the proposed RegGNN network is more stable to changes in the parameter $k$. Calculating for each architecture the average of the standard deviations (std) of the mean absolute error\footnote{The standard deviations include variations over different $k$.}  (see the results table)  over all feature extraction methods, we first note that the averages for RegGNN are 0.455, 0.145, 0.877, 0.369 for NT (FIQ), NT (VIQ), ASD (FIQ) and ASD (VIQ), respectively. While RegGNN therefore shows small variation with respect to $k$, CPM is highly sensitive to the changes of this parameter with averages of 2.901, 2.613, 1.803 and 2.754 respectively. This is approximately a 2 to 10 fold increase in variability. Consequently, RegGNN can better capture the brain graph structure, whereas CPM treats graphs as flattened vectors without preserving their topological features.}

    Improvements to the performance of CPM are not statistically significant ($p=0.87$ for ASD (FIQ), $p=0.15$ for ASD (VIQ)). Similarly, we observed that improvements to the performance of RegGNN are only statistically significant in NT (VIQ) task ($p<0.05$ for NT (VIQ), $p=0.98$ for NT (FIQ), $p=0.64$ for ASD (FIQ)). The performance increases for PNA models are more consistent, as PNA-S improved significantly in three out of four tasks ($p<0.01$ for NT (FIQ), $p=0.11$ for NT (VIQ), $p<0.05$ for ASD (FIQ), $p<0.01$ for ASD (VIQ)), and PNA-V improved significantly in two out of four tasks ($p<0.05$ for NT (FIQ), $p=0.21$ for NT (VIQ), $p=0.11$ for ASD (FIQ), $p<0.01$ for ASD (VIQ)).

    Within the sample selection pipelines, the best performing methods always utilize the Riemannian geometric structure of the SPD space with respect to MAE, apart from PNA-V results for the NT (VIQ) task. In the majority of cases, the methods that rely on tangent matrices perform best with the vectorized version of the whole tangent matrix being the best method in NT (VIQ) and ASD (FIQ). We also see that the three centrality measures and concatenated versions perform well consistently. Our results do not reveal any finer pattern among the sample selection measures, but we can conclude that using Riemannian structure of connectomes to estimate their predictive power outperforms methods that do not leverage these geometric properties. Thus, other metrics should also be considered when deciding for one. The tangent matrix method seems to perform very well but is also the most time consuming since no dimension reduction is performed. On the contrary, computing centrality measures significantly reduces the size of the matrices which speeds up the process sufficiently. In our experiments, we observed that training linear regression models using tangent matrices took up to 16 times more time compared to training models using centrality measures. To understand the latter methods and how they work better, it would be helpful to analyze their behavior on the tangent matrices mathematically. In contrast to their use for adjacency matrices of graphs, this is, to the best of our knowledge, unknown. It is thus an interesting venue for future work.
    
    An important advantage of using sample selection in training graph neural networks is the decrease in the computational power needed for the training process. Using the computational power more efficiently leads to shorter training times on fixed amount of data, which opens up opportunities to train more complex models on more data or in shorter amounts of time. While the exact time required for sample selection is heavily dependent on the hardware used and varies based on the model architecture, number of epochs in training, number of training samples, and the number $k$ of samples to select, our observations during the experiments show that sample selection reduces the training time by $20\%$ on average. Therefore, usage of our sample selection pipeline can enable the use of deeper neural network architectures on connectomes and provides a topic of interest in future work. 
    
    \majorrev{So far, we evaluated our method on a young population; however, our RegGNN demonstrated its generalizability by the utilized cross-validation strategy and across both NC and ASD brain connectivity datasets. The proposed model can be easily used to map a particular brain connectivity population (e.g., elderly population) to target scores to predict. To proliferate replication studies on other cohorts, we publicly shared our RegGNN source code\footnote{\url{https://github.com/basiralab/RegGNN}}.}

\textbf{Explainability and biomarker discovery.} 
    \majorrev{In order to identify the brain regions of interest that influence the prediction most, we extracted for each of the four tasks the learned weights of the RegGNN utilizing the best-performing sample selection method. The weights came from the fully connected layer that maps a 116-dimensional vector to the output score. (Remember that the input vector represents the learned embedding of the graph.) Thanks to the end-to-end network training, the backprogation process as well as our network design (which preserves the structure of the connectome in both the first and second layer), the learned weights in the final fully connected layer quantified the importance of its nodes in the target prediction task. Hence, a node with a higher weight in the fully connected layer is more influential for the prediction of the output score.}
    
    In Fig.~\ref{fig:brain_regions}, we show the regions of interest with the 3 highest weights averaged over $k = {2, \ldots, 15}$; underlying is the AAL parcellation atlas~\citep{TzourioMazoyer:2002}.\footnote{The brain networks were visualized with the \href{http://www.nitrc.org/projects/bnv/}{BrainNet Viewer} \citep{Xia:2013}.}
    For the FIQ prediction task in the NT cohort, we see that the left superior dorsal frontal gyrus (SFGdor.L), right superior frontal medial gyrus (SFGmed.R), and right cerebellum 6 (CRBL6.R) have the highest weights. For the VIQ prediction task in the same cohort, left hippocampus (HIP.L), left heschl gyrus (HES.L), and left cuneus (CUN.L) possess the highest weights.
    In the ASD cohort, the highest weights for FIQ prediction are left insula (INS.L), left calcarine cortex (CAL.L), and right pallidum (PAL.R), while the highest weights for VIQ prediction have the left superior frontal medial gyrus (SFGmed.L) left middle occipital gyrus (MOG.L), and left cuneus (CUN.L).
    
    According to our results, the more important regions of interest in IQ prediction lie in the left hemisphere of the brain. 
    Our findings are in line with other studies, that found that the insula shows greater activity in various cognitive tasks~\citep{Critchely:2000} and that the surface areal change in the left cuneus correlates strongly with full IQ, especially in perceptual tasks in young adults with very low birth weight~\citep{Skranes:2013}. Furthermore, we observe that the left cuneus was influential in predicting VIQ in both cohorts.
    Finally, as \citep{Dryburgh:2020}, our experiments indicate that the middle frontal gyrus is a significant region in IQ prediction. 
    
    \majorrev{A highly interesting question for future work is to investigate why the sample selection method improves the prediction, i.e, why there seem to be clusters within the data that can be represented by central samples. This is a challenging question that most likely requires the development of new analytical tools. Nevertheless, we think that it will be worth the effort as common structures and connections between these central samples could give us a lot more insights into the interplay between the connectivity structure of the brain and cognitive ability.}

\section{Conclusion}
    In this work, we applied RegGNN, a new graph neural network, to connectome data of neurotypical subjects and subjects with autism spectrum disorder to predict full scale and verbal intelligence quotients. We trained it using a novel sample selection method, which tries to identify samples within the training set that are expected to better predict the cognitive scores of new subjects. This enabled us to train the network with only 15 samples or less, while the testing performance was on par or even better than state-of-the-art methods for cognitive score prediction from connectomes. Both the sample selection and RegGNN are easy to implement in open access software and can be used in clinical practice.

\section*{Declarations}
{\small
\textbf{Funding} This work was funded by generous grants from the European H2020 Marie Sklodowska-Curie action (grant no. 101003403, \url{http://basira-lab.com/normnets/}) to I.R. and the Scientific and Technological Research Council of Turkey to I.R. under the TUBITAK 2232 Fellowship for Outstanding Researchers (no. 118C288, \url{http://basira-lab.com/reprime/}). However, all scientific contributions made in this project are owned and approved solely by the authors. M.A.G is funded by the same TUBITAK 2232 Fellowship. M.H. is funded by the Deutsche Forschungsgemeinschaft (DFG, German Research Foundation) under Germany's Excellence Strategy ? The Berlin Mathematics Research Center MATH+ (EXC-2046/1, project ID: 390685689).
\bigskip 

\noindent \textbf{Conflicts of interest/Competing interests} The authors declare that they have no conflict of interest.
\bigskip

\noindent \textbf{Ethics approval} Not applicable.
\bigskip

\noindent \textbf{Consent to participate} Not applicable.
\bigskip

\noindent \textbf{Consent for publication} All authors agreed to the publication of this article.
\bigskip

\noindent \textbf{Availability of data and material}
The ABIDE data that was used in this work is available online; the link is given in the text.
\bigskip

\noindent \textbf{Code availability}  
All methods were implemented in Python. Our RegGNN code is available at \\ \url{https://github.com/basiralab/RegGNN}.
\bigskip

\noindent \textbf{Author Contributions} Author contributions included designing the GNN and sample selection method as well as the experimental setup (all authors), implementation of the methods and experiments (M.A.D. and M.A.G.), writing the manuscript (M.H.) and revising it critically for important intellectual content (all authors), and approval of final version to be published
and agreement to be accountable for the integrity and accuracy of all aspects of the work (all
authors).

}

\appendix
    \section{Topological Centrality Measures}\label{Sec:AppA}
        \majorrev{Let $\mathbf{A}$ be the (weighted) adjacency matrix of $G$, $V$ the set of vertices of $G$, and $v \in V$.\footnote{With a slight abuse of notation, we identify nodes $v \in V$ and the integers we assign to them in order to construct $A$, e.g., we write $A_{vw}$ for the entry that corresponds to the edge between nodes $v$ and $w$.} The \emph{degree centrality $D(v)$ of $v$} is defined by 
        \begin{equation*} \label{eq:node_degree}
        D(v) := \sum_{\substack{w \in V\\w\ne v}} \mathbf{A}_{vw},
        \end{equation*}
        i.e., it assigns to each node its weighted sum of neighbors.}
        
        \majorrev{Let $x$ be the unit norm eigenvector of $A$ that corresponds to the largest eigenvalue $\lambda_1$ and has only non-negative entries.
        The \emph{eigenvector centrality $E(v)$ of $v$} is the $v$-th entry of $x$; that is,
        \begin{align*}
        &E(v) := \frac{1}{\lambda_1} \sum_{w \in V} \mathbf{A}_{vw} x_w, \label{eq:eigenvector_centrality}\\ 
        &\text{s.t. } \|x\|_2 = 1 \text{ and } x_w \ge 0 \text{ for all } w \in V. \nonumber
        \end{align*}
        It measures, in a relative sense, how influential a node is in the network.
        Intuitively, a high score means that a node has many neighbors that themselves have high eigenvector centrality scores.}
        
        \majorrev{Let $l_{vw}$ be the length of the shortest path between two nodes $v$ and $w$, and $n = |V|$.
        The \emph{closeness centrality $C(v)$ of $v$} is defined by
        \begin{equation*} \label{eq:closeness_centrality}
                C(v) := \frac{n-1}{\sum_{\substack{w \in V\\w\ne v}} l_{vw}},
        \end{equation*}
        i.e., as the inverse of the average distance of $v$ to all other nodes.}


\bibliographystyle{apalike}      
\bibliography{biblio}   

\end{document}